\documentclass[10pt,journal,compsoc]{IEEEtran}
\usepackage{epsfig}
\usepackage{graphicx}
\usepackage{amsmath}
\usepackage{amssymb}
\usepackage{multirow}
\usepackage{makecell}
\usepackage{tabulary}
\usepackage{stackengine}

\usepackage{url}
\usepackage[pagebackref=true,breaklinks=true,letterpaper=true,colorlinks,bookmarks=false]{hyperref}

\ifCLASSOPTIONcompsoc
  \usepackage[nocompress]{cite}
\else
  \usepackage{cite}
\fi
\ifCLASSINFOpdf
\else
\fi
\begin{document}
%
\title{Dual Vision Transformer}
%
%
%
%

\author{
Ting~Yao,~\IEEEmembership{Senior Member,~IEEE},
Yehao~Li,
Yingwei~Pan,~\IEEEmembership{Member,~IEEE},
Yu~Wang,\\
Xiao-Ping Zhang,~\IEEEmembership{Fellow,~IEEE},
and~Tao~Mei,~\IEEEmembership{Fellow,~IEEE}
\IEEEcompsocitemizethanks{\IEEEcompsocthanksitem
Ting~Yao, Yehao~Li, Yingwei~Pan, Yu~Wang, and Tao~Mei are with JD Explore Academy, Beijing, China (e-mail: tingyao.ustc@gmail.com; yehaoli.sysu@gmail.com; panyw.ustc@gmail.com; feather1014@gmail.com; tmei@jd.com).}
\IEEEcompsocitemizethanks{\IEEEcompsocthanksitem Xiao-Ping Zhang is with Ryerson University, Canada (email: xzhang@ee.ryerson.ca).}
\IEEEcompsocitemizethanks{\IEEEcompsocthanksitem Yingwei Pan is the corresponding author.}
\IEEEcompsocitemizethanks{\IEEEcompsocthanksitem This work was supported by the National Key R\&D Program of China under Grant No. 2020AAA0108600.}
\thanks{This work has been submitted to the IEEE for possible publication. Copyright may be transferred without notice, after which this version may no longer be accessible.}
}

\IEEEtitleabstractindextext{%
\begin{abstract}
Prior works have proposed several strategies to reduce the computational cost of self-attention mechanism. Many of these works consider decomposing the self-attention procedure into regional and local feature extraction procedures that each incurs a much smaller computational complexity. However, regional information is typically only achieved at the expense of undesirable information lost owing to down-sampling. In this paper, we propose a novel Transformer architecture that aims to mitigate the cost issue, named Dual Vision Transformer (Dual-ViT). The new architecture incorporates a critical semantic pathway that can more efficiently compress token vectors into global semantics with reduced order of complexity. Such compressed global semantics then serve as useful prior information in learning finer pixel level details, through another constructed pixel pathway. The semantic pathway and pixel pathway are then integrated together and are jointly trained, spreading the enhanced self-attention information in parallel through both of the pathways. Dual-ViT is henceforth able to reduce the computational complexity without compromising much accuracy. We empirically demonstrate that Dual-ViT provides superior accuracy than SOTA Transformer architectures with reduced training complexity. Source code is available at \url{https://github.com/YehLi/ImageNetModel}.
\end{abstract}

}

\maketitle

\IEEEdisplaynontitleabstractindextext

%
\IEEEpeerreviewmaketitle

\IEEEraisesectionheading{\section{Introduction}\label{sec:introduction}}

%
%
%
%

\begin{figure*}[!tb]
\centering {\includegraphics[width=1\textwidth]{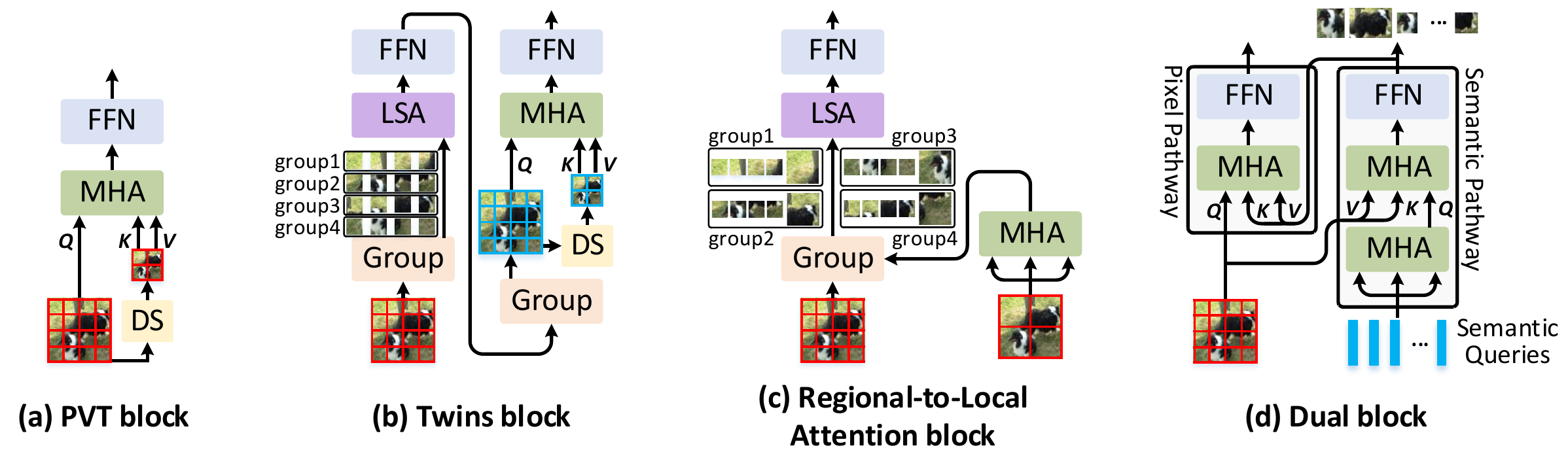}}
\vspace{-0.3in}
\caption{\small Illustration of (a) Pyramid Vision Transformer block (PVT), (b) Twins block that combines locally-grouped self-attention and spatial reduction attention, (c) Regional-to-Local Attention block in RegionViT, and (d) our proposed Dual block in Dual-ViT. DS: down-sampling operation; MSA: multi-head self-attention; FFN: feed-forward layer; LSA: locally-grouped self-attention. Layer normalization and residual connection are omitted for simplicity.}
\label{fig:fig1}
\vspace{-0.1in}
\end{figure*}

\IEEEPARstart{R}{ecently,} Transformer architectures \cite{cao2020global,dosovitskiy2020image,han2022survey,liu2021swin,vaswani2017attention,xiao2022image} have shown great success in revolutionizing deep learning applications, including both natural language processing and computer vision tasks. Unfortunately, as Transformers typically rely on dense self-attention computations, the training of such architectures are usually slow for high-resolution inputs. As transformer technologies can often provide superior performance than its counterparts, this complexity issue gradually becomes a bottleneck constraining the progress of such powerful architecture.

Self-attention procedure forms the main burden in such complexity issue, as each the representation of each token is updated by attending to all the tokens. The recent works then turn their focus on investigating the complexity issue by providing various alternative solutions over the standard self-attention. Many consider combining self-attention with down-sampling to effectively replace the original standard attention. Such way naturally enables the exploration of regional semantic information, which further boosts the learning/extraction of local finer features. Take for instance, PVT \cite{wang2021pyramid,wang2021pvtv2} proposes linear spatial reduction attention (SRA) that reduces the spatial scale of keys and values with the down-sampling operations (e.g., average pooling or strided convolution), as showed in Figure \ref{fig:fig1}(a). Twins \cite{chu2021twins} (Figure \ref{fig:fig1}(b)) adds the additional locally-grouped self-attention layer before SRA to further enhance the representation via intra-region interaction. RegionViT \cite{chen2021regionvit} (Figure \ref{fig:fig1}(c)) decomposes the original attention through regional and local self-attentions. However, since the above methods heavily rely on the down-sampling of the feature maps into regions, evident performance drop has been observed while the total computational cost is effectively saved.

Among these various combination strategies, there has been little attempt to investigate dependency between global semantics and finer pixel level features in terms of complexity reduction. In this paper, we consider decomposing the training into global semantics and finer features attentions via the proposed Dual-ViT. The incentive is to extract global semantic information (i.e., parametric semantic queries) that can serve as rich prior information to assist finer local feature extraction in a new two-pathway design. Our unique decomposition and integration of both the global semantics and local features allow for effective reduction of the involved number of tokens in multi-head attention, thus saving the computational complexity in contrast to the standard attention counterparts. Especially, as showed in Figure \ref{fig:fig1}(d), Dual-ViT consists of two special pathways respectively called ``semantic pathway'' and ``pixel pathway''. The local pixel level feature extraction through the constructed ``pixel pathway'' is imposed by strong dependency on the compressed global prior out of the ``semantic pathway''. Since gradients pass through both the semantic pathway and the pixel pathway, the Dual-ViT training procedure can henceforth efficiently both compensate the information loss on global feature compression, and simultaneously reduce the difficulty of finer local feature extraction. Both the former and latter procedure can in parallel significantly reduce the computation cost, owing to smaller attention sizes and the imposed dependency between the two pathways.

In summary, we have made the following contributions in this paper:
\begin{enumerate}
\item{} We propose a novel Transformer architecture called Dual Vision Transformer (Dual-ViT). As its name implies, Dual-ViT network includes two pathways that are tailored respectively to extract more holistic global view of semantic feature of the input, and another pixel pathway that focuses on the learning of finer local features.
\item{} Dual-ViT takes into account the dependency between the global semantics and local features along the two pathways, with the goal to ease the training with reduced token sizes and smaller attentions.
\item{} Dual-ViT achieves 85.7\% top-1 accuracy on ImageNet with only 41.1\% FLOPs and 37.8\% parameters compared to VOLO \cite{yuan2021volo}. In regard of object detection and instance segmentation, Dual-ViT also improves PVT \cite{wang2021pvtv2} in term of mAP by more than 1.2\% and 0.9\% on COCO, with 48.0\% less parameters.
\end{enumerate}

\section{Related Work}
\subsection{Convolutional Neural Networks Backbones}
Convolutional Neural Networks (CNNs) have dominated the modeling of deep architectures in computer vision field in the last decade. Taking the inspiration from the breakthrough achieved with AlexNet \cite{krizhevsky2012imagenet} on ImageNet benchmark, a series of powerful CNN backbones have been established subsequently. For example, VGG \cite{simonyan2014very} is one of the early attempts that increases the depth to strengthen network capability, and ResNets \cite{he2016deep} further introduces skip connections into CNN block, enabling the training of very deep networks (152 layers). In addition to make the network deeper, multiple paths strategy is another direction for facilitating visual representation learning. In between, GoogLeNet \cite{szegedy2015going} and InceptionNet \cite{szegedy2016rethinking} introduce the split-transform-merge strategy that integrates multiple kernel paths into a single CNN block. ResNeXt \cite{xie2017aggregated} goes beyond ResNet by designing a homogeneous and multi-branch architecture that increases accuracy while maintains the computational complexity. In contrast to InceptionNet that employs different kernel sizes in a CNN block, elastic \cite{wang2019elastic} leverages different resolutions branches to learn the multi-scale representation. Res2Net \cite{gao2019res2net} improves the multi-scale ability at a more granular level with the hierarchical residual-like connections. HRNet \cite{wang2020deep} maintains high resolution through the whole process and fuses multi-resolution representations repeatedly, yielding semantically strong representations in both high and low resolutions. SKNet \cite{li2019selective} dynamically fuses the multiple paths with different kernel sizes via softmax attention mechanism. Recently, EfficientNet \cite{tan2019efficientnet} designs a new scaling method that seeks a better balance between network width, depth, and resolution.

\begin{figure*}[!tb]
\centering {\includegraphics[width=1\textwidth]{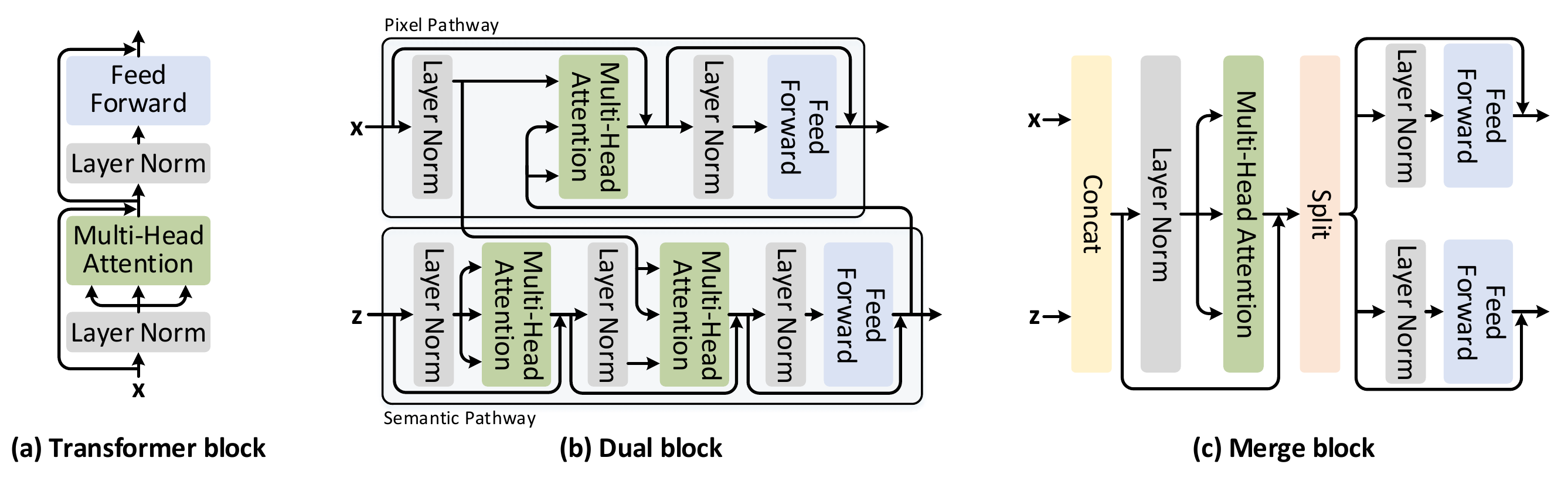}}
\vspace{-0.36in}
\caption{\small The detailed architectures of (a) Transformer block, (b) Dual block, and (c) Merge block.}
\label{fig:framework}
\vspace{-0.2in}
\end{figure*}

\subsection{Vision Transformers}
Sparked by the success of Transformer \cite{vaswani2017attention} in natural language processing field, numerous Transformer-based backbones start to emerge. Early attempts replace the convolutional layers with self-attention network in ResNet architecture \cite{hu2019local,ramachandran2019stand,zhao2020exploring,li2022contextual}. These techniques trigger interaction within a local window via self-attention, which achieve better performances while the FLOPs are reduced. Recently, Vision Transformer (ViT) \cite{dosovitskiy2020image} proposes a pure Transformer backbone that operates over the image patch sequences for image recognition. Unlike the conventional CNNs, ViT does not use any pooling layer and instead leverages a relatively lower spatial resolution for all layers, which is not suitable for pixel-level dense prediction tasks that require high-resolution outputs. PiT \cite{heo2021rethinking} remoulds ViT into a multi-scale version by using the common design principle in CNNs, i.e.,  starting from a large spatial size and then decreasing the spatial resolution gradually via pooling layers. However, PiT suffers from heavy computational cost of self-attention at earlier stages with high-resolution inputs. To tackle this issue, Swin \cite{liu2021swin} designs the shifted windowing scheme by restricting self-attention within local windows while allowing for cross-window connections. PVT \cite{wang2021pvtv2,wang2021pyramid} reduces the spatial scale of keys and values with the down-sampling operations. Multiscale Vision Transformer \cite{fan2021multiscale} employs pooling kernels over query,keys and values for spatial reduction. Twins \cite{chu2021twins} combines locally-grouped self-attention and sub-sampled self-attention to exploit intra- and inter-regional interaction. RegionViT \cite{chen2021regionvit} presents regional-to-local attention to alleviate the overhead of standard self-attention by interleaving regional and local self-attention in the opposite order of Twins.

Our Dual-ViT is also a type of multi-scale ViT backbone. In comparison to existing multi-scale ViTs that heavily rely on local self-attention within a local window or down-sampling operations, Dual-ViT decomposes the modeling of self-attention into the learning of global semantics and finer features in two pathways. Both of semantic tokens and input feature are further integrated to spread the enhanced self-attention information in parallel. Such unique design of decomposition and integration not only effectively reduces the number of tokens in self-attention learning, but also imposes the interactions between the two pathways, leading to better accuracy and latency trade-off.

\section{Method}
This section begins with a brief review of the conventional multi-head self-attention block adopted in existing ViTs, and analyzes how they scale down the self-attention computational cost. We next propose a new principled Transformer structure, namely Dual Vision Transformer (Dual-ViT). Our launching point is to upgrade typical Transformer structure with a particular two-pathway design and trigger the dependency between the global semantics and local features to enhance self-attention learning.

Specifically, Dual-ViT is composed of four stages where the resolutions of the feature maps in each stage shrink gradually as in \cite{liu2021swin}. In the first two stages with high-resolution inputs, Dual-ViT employs the new Dual blocks that consist of two pathways: ($i$) pixel pathway that captures fine-grained information by refining input feature at pixel level, and ($ii$) semantic pathway that abstracts high-level semantic tokens at global level. The semantic pathway is slightly deeper (with more operations) but contains fewer semantic tokens abstracted from pixels, and the pixel pathway treats these global semantics as prior to learn finer pixel level details. Such design conveniently encodes the dependency of finer information on holistic semantics, and meanwhile reduces the computational cost of multi-head self-attention over high-resolution inputs. The outputs of the two pathways are merged together and further fed into multi-head self-attention in the last two stages.

\subsection{Conventional Transformer}
Conventional Transformer architectures (e.g., \cite{dosovitskiy2020image,touvron2021training}) often rely on multi-head self-attention that captures long-range dependencies among inputs. Here we present a general formulation of multi-head self-attention (see Figure \ref{fig:framework}(a)). Transformer block consists of a multi-head attention layer (${\bf{MHA}}$) and a feed-forward layer (${\bf{FFN}}$). Layer normalization (${\bf{LN}}$) is applied before each block. Technically, given the input feature $x_l \in {{\mathbb{R}}^{n \times d}}$ ($n = H \times W$: the number of tokens, $H$/$W$/$d$: height/width/channel number), the $l$-th conventional Transformer block operates as follows:
\begin{equation}
\label{eq:trb}
\begin{aligned}
&{{x^n_l}={\bf{LN}}({x_l})},\\
&{{x'_l} = {\bf{MHA}}({x^n_l, x^n_l, x^n_l}) + {x_l}},\\
&{{\bf{MHA}}(q, k, v) = {\bf{Concat}}({head}_{1},...,{head}_{h}){W^O}},\\
&{{head}_{i} = {\bf{Attention}}(qW_i^Q, kW_i^K, vW_i^V)}, \\
&{ {\bf{Attention}}({\bf{Q}}, {\bf{K}}, {\bf{V}}) = {\bf{softmax}}(\frac{{{\bf{Q}}{{\bf{K}}^T}}}{{\sqrt {d_h} }}){\bf{V}}},\\
&{{x_{l + 1}} = {\bf{FFN}}({\bf{LN}}({x'_l})) + {x'_l}},\\
\end{aligned}
\end{equation}
where ${\bf{Concat}}(\cdot)$ is the concatenation operation, $W_i^Q$, $W_i^K$, $W_i^V$, $W^O$ are weight matrices, and $d_h$ is the dimension of each head. According to the formulation in Eq.(\ref{eq:trb}), the computation cost of ${\bf{MHA}}$ for input feature $x_l$ is $\mathcal{O}(n^2d)$, which scales quadratically w.r.t. the number of tokens. As a result, such design will inevitably result in huge computation cost when processing high-resolution inputs. In order to alleviate such heavy computation cost of ${\bf{MHA}}$, existing techniques \cite{dosovitskiy2020image,touvron2021training,yuan2021volo} enlarge the patch size to produce fewer tokens, leading to lower-resolution feature maps. Nevertheless, it is not trivial to directly employ these backbones in dense prediction tasks (e.g., object detection and semantic segmentation) that require high-resolution feature maps. Other works \cite{chen2021regionvit,chu2021twins,liu2021swin} restrict the attention scope into a local window and thus can achieve only linear computational complexity with respect to the input resolution. However, the limited receptive field of local window adversely hinders the modeling of global dependencies, resulting in a sub-optimal solution. The advances \cite{fan2021multiscale,guo2021cmt,wang2021pvtv2} introduce spatial reduction attention by using the down-sampling operations (e.g., average pooling in \cite{wang2021pvtv2} or pooling kernels in \cite{fan2021multiscale,guo2021cmt}) over keys/values to reduce computation cost. But these pooling-based operations inevitably incur information loss and thus hurt the performances.

\subsection{Dual Block}
To mitigate the aforementioned issues, we design a principled self-attention block tailored to high-resolution inputs (i.e., at the first two stages), namely \textbf{Dual block}. This new design nicely introduces an additional pathway to ease self-attention learning. Figure \ref{fig:framework}(b) depicts the detailed architecture of Dual block. Specifically, Dual block contains two pathways: pixel pathway and semantic pathway. The semantic pathway summarizes the input feature map into semantic tokens. After that, the pixel pathway takes these semantic tokens as rich prior in the form of keys/values, and performs multi-head attention to refine input feature map via cross-attention. In terms of complexity, since the semantic pathway contains much fewer tokens than that in pixel pathway, the computation cost is reduced to $\mathcal{O}(nmd + m^2d)$, where $m$ is the number of semantic tokens.

Formally, given the input feature $x_l$ of the $l$-th Dual block, we augment with additional parametric semantic queries $z_l \in {{\mathbb{R}}^{m \times d}}$. The semantic pathway first contextually encodes the semantic queries via self-attention, and then extracts the semantic tokens by exploiting the interaction between refined semantic queries and the input feature $x_l$ via cross-attention, followed by a feed-forward layer. This operation performs as follows:
\begin{equation}
\label{eq:semantic}
\begin{aligned}
&{{x^n_l}={\bf{LN}}({x_l}), {z^n_l}={\bf{LN}}({z_l})}, \\
&{{z'_l} = {\bf{MHA}}({z^n_l, z^n_l, z^n_l}) + {z_l}},\\
&{{\tilde z_l} = {\bf{MHA}}({{\bf{LN}}({z'_l}), x^n_l, x^n_l}) + {z'_l}},\\
&{{z_{l + 1}} = {\bf{FFN}}({\bf{LN}}({\tilde z_l})) + {\tilde z_l}}.\\
\end{aligned}
\end{equation}
The semantic tokens $z_{l + 1}$ are fed into the pixel pathway and serve as prior information of high-level semantics. Meanwhile, we treat the semantic tokens as the enhanced semantic queries and feed them into the semantic pathway of next Dual block.

The pixel pathway plays a similar role as conventional Transformer block, except that it additionally takes the semantic tokens derived from semantic pathway as prior to refine input feature via cross-attention. More specifically, the pixel pathway treats the semantic tokens ${z_{l + 1}}$ as keys/values, and performs cross-attention as follows:
\begin{equation}
\label{eq:pixel}
\begin{aligned}
&{{x^n_l}={\bf{LN}}({x_l}), {z^n_{l+1}}={\bf{LN}}({z_{l+1}})}, \\
&{{x'_l} = {\bf{MHA}}({x^n_l, z^n_{l+1}, z^n_{l+1}}) + {x_l}},\\
&{{x_{l + 1}} = {\bf{FFN}}({\bf{LN}}({x'_l})) + {x'_l}}.\\
\end{aligned}
\end{equation}

Considering that the gradients back propagate through both two pathways, Dual block is thus able to simultaneously compensate the information loss on global feature compression through pixel-to-semantic interaction, and reduce the difficulty of finer local feature extraction with the global prior via semantic-to-pixel interaction.

\begin{table*}[!tb]\small
\centering
\caption{Detailed architecture specifications for three variants of our Dual-ViT with different model sizes, i.e., Dual-ViT-S (Small size), Dual-ViT-B (Base size), and Dual-ViT-L (Large size). $HD_i$, $C_i$, and $E^x$/$E^z$ represents the head number, channel dimension, and the expansion ratio of feed-forward layer for the tokens from pixel/semantic pathway in stage $i$, respectively.}
\vspace{-0.10in}
\begin{tabular}{ccccc}
\Xhline{2\arrayrulewidth}
Specification & & Dual-ViT-S & Dual-ViT-B & Dual-ViT-L \\ \hline
Patch Embedding & & 4 $\times$ 4 & 4 $\times$ 4 & 4 $\times$ 4 \\ \hline
Stage 1 ($\frac{H}{4} \times \frac{W}{4}$)~~~~
        & $\begin{array}{c} {\bf{Dual}} \\ {\bf{Block}} \end{array}$
        & $\left[ \begin{array}{c}  HD_1=2 \\ C_1=64 \\ E^x_1=8 \\ E^z_1=4 \end{array} \right] \!\times\! 3$ ~
        & $\left[ \begin{array}{c}  HD_1=2 \\ C_1=64 \\ E^x_1=8 \\ E^z_1=4 \end{array} \right] \!\times\! 3$ ~
        & $\left[ \begin{array}{c}  HD_1=3 \\ C_1=96 \\ E^x_1=8 \\ E^z_1=4 \end{array} \right] \!\times\! 3$
        \\ \hline
Patch Embedding & & 2 $\times$ 2 & 2 $\times$ 2 & 2 $\times$ 2 \\ \hline
Stage 2 ($\frac{H}{8} \times \frac{W}{8}$)~~~~
        & $\begin{array}{c} {\bf{Dual}} \\ {\bf{Block}} \end{array}$
        & $\left[ \begin{array}{c}  HD_2=4 \\  C_2=128 \\ E^x_2=8 \\ E^z_2=4 \end{array} \right] \!\times\! 4$ ~
        & $\left[ \begin{array}{c}  HD_2=4 \\  C_2=128 \\ E^x_2=8 \\ E^z_2=4 \end{array} \right] \!\times\! 4$ ~
        & $\left[ \begin{array}{c}  HD_2=6 \\  C_2=192 \\ E^x_2=8 \\ E^z_2=4 \end{array} \right] \!\times\! 6$
        \\ \hline
Patch Embedding & & 2 $\times$ 2 & 2 $\times$ 2 & 2 $\times$ 2 \\ \hline
Stage 3 ($\frac{H}{16} \times \frac{W}{16}$)~~~~
        & $\begin{array}{c} {\bf{Merge}} \\ {\bf{Block}} \end{array}$
        & $\left[ \begin{array}{c}  HD_3=10 \\ C_3=320 \\ E^x_3=4 \\ E^z_3=2 \end{array} \right] \!\times\! 6$ ~
        & $\left[ \begin{array}{c}  HD_3=10 \\ C_3=320 \\ E^x_3=4 \\ E^z_3=2 \end{array} \right] \!\times\! 15$ ~
        & $\left[ \begin{array}{c}  HD_3=12 \\ C_3=384 \\ E^x_3=4 \\ E^z_3=2 \end{array} \right] \!\times\! 21$
        \\ \hline
Patch Embedding & & 2 $\times$ 2 & 2 $\times$ 2 & 2 $\times$ 2 \\ \hline
Stage 4 ($\frac{H}{32} \times \frac{W}{32}$)~~~~
        & $\begin{array}{c} {\bf{Merge}} \\ {\bf{Block}} \end{array}$
        & $\left[ \begin{array}{c}  HD_4=14 \\ C_4=448 \\ E^x_4=3 \\ E^z_4=2 \end{array} \right] \!\times\! 3$ ~
        & $\left[ \begin{array}{c}  HD_4=16 \\ C_4=512 \\ E^x_4=3 \\ E^z_4=2 \end{array} \right] \!\times\! 3$ ~
        & $\left[ \begin{array}{c}  HD_4=16 \\ C_4=512 \\ E^x_4=3 \\ E^z_4=2 \end{array} \right] \!\times\! 3$
        \\ \Xhline{2\arrayrulewidth}
\end{tabular}
\label{table:architecture}
\vspace{-0.2in}
\end{table*}

\subsection{Merge Block}
Recall that the dual blocks in the first two stages exploit the inter-interaction between two pathways, while leaving the intra-interaction among local tokens within pixel pathway unexploited due to huge complexity for high-resolution inputs. To alleviate this issue, we present a simple yet effective design of self-attention block (namely \textbf{Merge block}) to
perform self-attention over concatenated semantic and local tokens in the last two stages (with low-resolution inputs), thereby enabling the intra-interaction among local tokens.
Figure \ref{fig:framework}(c) depicts the architecture of Merge block. Specifically, we directly merge the output tokens from both pathways and feed them into multi-head self-attention layers. Owing to the fact that the tokens from the two pathways convey different information, two separate feed-forward layers are employed for each pathway in Merge block:
\begin{equation}
\label{eq:merge}
\begin{aligned}
&{y^n_l = {\bf{LN}}([~{x_l}||{z_l}~])}, \\
&{{x'_l}, {z'_l} = {\bf{MHA}}({y^n_l, y^n_l, y^n_l}) + {[~{x_l}||{z_l}~]}},\\
&{x_{l+1} = {\bf{FFN^x}}({\bf{LN}}({x'_l})) + {x'_l}},\\
&{z_{l+1} = {\bf{FFN^z}}({\bf{LN}}({z'_l})) + {z'_l}},\\
\end{aligned}
\end{equation}
where $[~||~]$ denotes the tensor concatenation, ${\bf{FFN^x}}$ and ${\bf{FFN^z}}$ are two different feed-forward layers. Finally, we employ global average pooling over the output tokens of two pathways to produce the final classification token.

\subsection{Dual Vision Transformer}
Our proposed Dual block and Merge block are essentially unified self-attention blocks. It is thus feasible to construct multi-scale ViT backbones by stacking these blocks. Following the basic configuration of existing multi-scale ViTs \cite{liu2021swin,wang2021pyramid}, the complete Dual-ViT contains four stages. The first two stages are comprised of a stack of Dual blocks while the last two stages consist of Merge blocks. According to the design principle of CNN architectures, a patch embedding layer is employed to increase the channel dimension and meanwhile shrink the spatial resolution at the beginning of each stage. In this work, we present three variants of Dual-ViT in different model sizes, i.e., Dual-ViT-S (Small size), Dual-ViT-B (Base size), and Dual-ViT-L (Large size). Note that Dual ViT-S/B/L shares similar model size and computational complexity with Swin-T/S/B \cite{liu2021swin}. Table \ref{table:architecture} details the architectures of all the three variants of Dual-ViT, where $HD_i$, $C_i$, and $E^x$/$E^z$ is the head number, channel dimension, and the expansion ratio of feed-forward layer for the tokens derived from pixel/semantic pathway in stage $i$.

\subsection{Differences between Our Dual-ViT and Previous Vision Transformers}
In this section, we discuss the detailed differences between our Dual-ViT and the precedent relevant ViT backbones.

\textbf{RegionViT} \cite{chen2021regionvit} designs the regional-to-local attention, which involves two kinds of tokens: the regional tokens with larger patch size and the local tokens with smaller patch size. RegionViT first performs regional self-attention over all regional tokens to learn global information. Then the local self-attention exchanges the information between each single regional token and its associated local tokens within a local window. Our Dual-ViT differs from this work in two aspects: First, the semantic tokens in Dual block are not constrained to be the uniform patches of equally size as in RegionViT, and thus are more flexible in encoding semantics; Second, the global semantic tokens act as holistic prior to compensate information loss in pixel pathway, whereas the regional tokens of RegionViT only interact with its associated local tokens within a local~window.

\textbf{Twins} \cite{chu2021twins} is composed of two types of attention operations: local and sub-sampled self-attention. Specifically, Twins divides the input tokens into several groups and performs local self-attention over each group within each local sub-window. In order to enable interactions between different sub-windows, Twins utilizes sub-sampled self-attention by down-sampling the feature map into regional tokens which act as down-sampled keys and values as in \cite{wang2021pvtv2}. The difference between Dual-ViT and Twins is that our semantic tokens are holistically summarized from the whole feature map, while Twins independently produces each regional token from a local sub-window.

\textbf{CrossViT} \cite{chen2021crossvit} contains two kinds of tokens with different patch sizes, which are fed into two Transformer encoders in two separate branches. Finally, CrossViT integrates the $CLS$ token of each branch with the output patch tokens of another branch. In contrast, our Dual-ViT triggers the interaction between the tokens from the pixel and semantic pathways within each Dual block, rather than encoding each branch with separate transformer encoder as in CrossViT. Moreover, Dual-ViT exchanges the information between multiple global-level semantic tokens and finer pixel-level features throughout the whole process, while CrossViT only interacts the final output features with a single compressed $CLS$ token, which may lack in the detailed semantics.

\section{Experiments}
We verify the merit of our Dual-ViT via various empirical evidences on multiple vision tasks, e.g., image recognition, object detection, instance segmentation, and semantic segmentation. In particular, we first use the most common image recognition benchmark (ImageNet \cite{deng2009imagenet}) to train the proposed Dual-ViT from scratch. Next, the pre-trained Dual-ViT is fine-tuned over COCO \cite{lin2014microsoft} and ADE20K \cite{zhou2019semantic} for the downstream tasks of object detection, instance segmentation, and semantic segmentation, aiming to evaluate the generalization capability of the pre-trained Dual-ViT.

\subsection{Datasets}
ImageNet is a large-scale image recognition benchmark and contains 1.28 million training images plus 50K validation images from 1,000 object categories. For dense prediction tasks of object detection and instance segmentation, we adopt the COCO dataset, which consists of $\sim$118K images (train2017) for training and 5K images (val2017) for evaluation. ADE20K is a common benchmark for semantic segmentation, and comprises 25K images from 150 semantic categories. The 25K images are divided into 20K, 2K, and 3K in training, validation, and test sets, respectively.

\begin{figure*}[!tb]
\qquad
\begin{tabular}[b]{c|cc|cc}
\hline
         &  Params  & GFLOPs  & Top-1 & Top-5 \\ \hline
  a      & 24.8M    & 5.4     & 83.9  & 96.7  \\
  b      & 24.4M    & 5.4     & 83.8  & 96.6  \\
  c      & 25.1M    & 5.4     & 84.0  & 96.7  \\
  d      & 25.1M    & 5.4     & \textbf{84.1}  & \textbf{96.8}  \\ \hline
\end{tabular}
~~~~~~~~~~~
\centering {\includegraphics[width=0.5\textwidth]{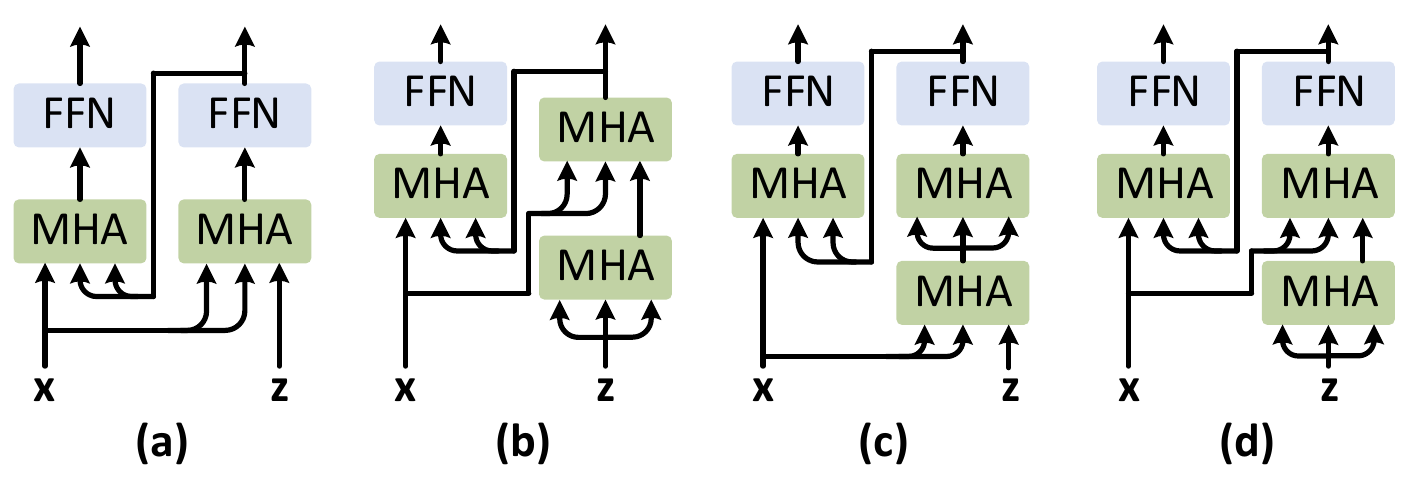}}
\vspace{-0.1in}
\caption{Ablation experiments by comparing different variants of Dual block: (a) and (b) removes the self-attention layer and feed-forward layer in the semantic pathway of Dual block, respectively; (c) exchanges the order of self-attention layer and cross-attention layer in the semantic pathway of Dual block (i.e., first performing cross-attention, and then employing self-attention); (d) the complete version of Dual block. The backbone network is fixed as Dual-ViT-S, and we only replace each Dual block in the first two stages with its variant for evaluation.}
\label{fig:ablation}
\vspace{-0.1in}
\end{figure*}

\subsection{Experimental Settings of Each Vision Task}

\textbf{Image Recognition} aims to classify each image into 1,000 object categories. In this task, we directly train our Dual-ViT on the training set of ImageNet, and the top-1 \& top-5 accuracies on the validation set are reported for evaluation. During training, we optimize the whole architecture with AdamW optimizer \cite{loshchilov2017decoupled} (momentum: 0.9) on 8 V100 GPUs. Each image is resized at typical resolution (254 $\times$ 254). In addition, we adopt several common data augmentation strategies in state-of-the-art vision backbones \cite{yuan2021volo}, e.g., CutOut \cite{zhong2020random}, RandAug \cite{cubuk2020randaugment}, and Token Labeling with MixToken \cite{jiang2021all}. The whole optimization spends 300 epochs with cosine decay learning rate scheduler \cite{loshchilov2016sgdr} and 10 epochs of linear warm-up. The batch size, learning rate and weight decay are set as 1,024, 0.001 and 0.05.

\textbf{Object Detection} task targets for localizing and recognizing the instances of objects within an image. Here we adopt five common and state-of-the-art object detectors (i.e., RetinaNet \cite{lin2017focal}, Cascade Mask R-CNN \cite{cai2018cascade}, ATSS \cite{zhang2020bridging}, GFL \cite{li2020generalized}, and Sparse RCNN \cite{sun2021sparse}) for this task. The CNN backbone in each object detector is directly replaced with our Dual-ViT. In particular, we first pre-train each vision backbone on ImageNet, and initialize all the newly included layers for object detection with Xavier \cite{glorot2010understanding}. Next, the whole detector architecture is trained on COCO train2017 set with the standard setting as in \cite{liu2021swin}. We evaluate the learnt detector on COCO val2017 set by reporting the Average Precision score ($AP$) at different IoU thresholds and for three different object sizes (i.e., small ($AP_S$), medium ($AP_M$), large ($AP_L$)). At the training stage, we set the batch size as 16, and leverage AdamW \cite{loshchilov2017decoupled} with 0.0001 initial learning rate and 0.05 weight decay. Each image is resized by maintaining the shorter side as 800 pixels and ensuring that the longer side do not exceed 1,333 pixels. More specifically, for RetinaNet, we use the 1 $\times$ training schedule (12 epochs). For Cascade Mask R-CNN, ATSS, GFL, and Sparse RCNN, we choose the 3 $\times$ schedule (36 epochs) with multi-scale strategy as in \cite{liu2021swin,wang2021pvtv2}, where the shorter side of each image is randomly resized within [480, 800] and the longer side is limited within 1,333 pixels.

\textbf{Instance Segmentation} task goes beyond object detection by localizing objects from bounding-box level to pixel level. The most common technique, i.e., Mask R-CNN \cite{he2017mask}, is adopted in this task. Similar to the task of object detection, we pre-train vision backbone on ImageNet, initialize the newly added layers with Xavier, and then fine-tune the whole architecture on COCO train2017. COCO val2017 is utilized to evaluate the learnt model, and both bounding box \& mask Average Precision scores ($AP^b$, $AP^m$) are reported at inference. Here we adopt the same training setups as in RetinaNet.

\textbf{Semantic Segmentation} aims to predict the semantic label of each pixel without instance-level discrimination. In the experiments, we adopt UPerNet \cite{xiao2018unified} as the basic architecture in this task and replace the CNN backbone with our pre-trained Dual-ViT. We train and evaluate UPerNet with the backbone of Dual-ViT on ADE20K dataset. The evaluation metric is mean IoU (mIoU) averaged over all classes.
At the training stage, the whole architecture is trained on 8 GPUs for 160K iterations via AdamW \cite{loshchilov2017decoupled} optimizer with linear learning rate decay scheduler and 1,500 iterations linear warmup. The batch size, initial learning rate and weight decay are set as 16, 0.00006 and 0.01, respectively. Each input image is cropped as 512 $\times$ 512. Several common data augmentation operations, i.e., random photometric distortion, random horizontal flipping, and random re-scaling within ratio range [0.5, 2.0] are adopted. All the hyperparameters and detection heads are set as in Swin \cite{liu2021swin} for fair comparison.

\begin{table*}[!tb]
\centering
\caption{Comparison results of our Dual-ViT with other state-of-the-art vision backbones on ImageNet for image recognition at typical resolution (254 $\times$ 254). We split all runs into 3 segments according to the model size (Small, Base, Large). $^{\dagger}$ denotes the using of additional Token Labeling objective via MixToken~\cite{jiang2021all}.}
\vspace{-0.1in}
\begin{tabular}{c|c|c|cc|c|c|c|cc}
\Xhline{2\arrayrulewidth}
Network       & Params & GFLOPs & Top-1 & Top-5 & Network      & Params & GFLOPs & Top-1 & Top-5 \\ \hline
\multicolumn{5}{c|} {Small} & \multicolumn{5}{c} {Base} \\ \hline
ResNet-50 \cite{he2016deep}    & 25.5M  & 4.1   & 78.3  & 94.3  & ResNet-101 \cite{he2016deep}  & 44.6M  & 7.9   & 80.0  & 95.0  \\
DeiT-S \cite{touvron2021training}       & 22.1M  & 4.6   & 79.9  & 95.0  & T2T-ViTt-19 \cite{yuan2021tokens}  & 39.2M  & 9.8   & 82.2  & 96.1     \\
Swin-T \cite{liu2021swin}       & 29.0M  & 4.5   & 81.2  & 95.5  & CvT-21 \cite{wu2021cvt}      & 32.0M  & 7.1   & 82.5  & -     \\
ConViT-S \cite{d2021convit}     & 27.8M  & 5.4   & 81.3  & 95.7  & CaiT-S24 \cite{touvron2021going}    & 46.9M  & 9.4   & 82.7  & -     \\
TNT-S \cite{han2021transformer}        & 23.8M  & 5.2   & 81.5  & 95.7  & CrossViT-18 \cite{chen2021crossvit} & 44.3M  & 9.5   & 82.8  & -     \\
SE-CoTNetD-50  \cite{li2022contextual} & 23.1M & 4.1 & 81.6 & 95.8      & Swin-S \cite{liu2021swin}      & 50.0M  & 8.7   & 83.2  & 96.2  \\
T2T-ViTt-14 \cite{yuan2021tokens}   & 21.5M  & 6.1   & 81.7  & 95.8     & Twins-SVT-B \cite{chu2021twins} & 56.1M    & 8.6   & 83.2  & 96.3   \\
CaiT-XS24 \cite{touvron2021going}    & 26.6M  & 5.4   & 81.8  & -       & SE-CoTNetD-101 \cite{li2022contextual} & 40.9M & 8.5   & 83.2 & 96.5 \\
PVTv2-B2 \cite{wang2021pvtv2}     & 25.4M  & 4.0   & 82.0  & 95.6       & PVTv2-B3 \cite{wang2021pvtv2}    & 45.2M  & 6.9   & 83.2  & 96.5    \\
CrossViT-15 \cite{chen2021crossvit}  & 28.2M  & 6.1   & 82.3  & -       & RegionViT-M+ \cite{chen2021regionvit} & 42.0M  & 7.9   & 83.4  & -     \\
RegionViT-S \cite{chen2021regionvit}  & 30.6M  & 5.3   & 82.6  & 96.1   & PVTv2-B4 \cite{wang2021pvtv2}     & 62.6M  & 10.1  & 83.6  & 96.7   \\
Dual-ViT-S    & 24.6M  & 4.8   & \textbf{83.4}  & \textbf{96.5}  & VOLO-D1$^{\dagger}$ \cite{yuan2021volo}    & 26.6M    & 6.8   & 84.2  & 96.8    \\
Dual-ViT-S$^{\dagger}$    & 25.1M  & 5.4   & \textbf{84.1}  & \textbf{96.8}  & Dual-ViT-B$^{\dagger}$  & 42.6M  & 9.3   & \textbf{85.2}  & \textbf{97.2}  \\ \hline
\multicolumn{10}{c} {Large} \\ \hline
ResNet-152 \cite{he2016deep}   & 60.2M  & 11.6  & 81.3  & 95.5            & Swin-B \cite{liu2021swin}      & 88.0M  & 15.4  & 83.5  & 96.5     \\
ResNeXt101 \cite{xie2017aggregated}   & 83.5M  & 15.6  & 81.5  & -        & Twins-SVT-L \cite{chu2021twins}  & 99.3M  & 15.1  & 83.7  & 96.5     \\
DeiT-B   \cite{touvron2021training}      & 86.6M  & 17.6  & 81.8  & 95.6  & RegionViT-B+ \cite{chen2021regionvit} & 73.8M  & 13.6  & 83.8  & -     \\
ResNeSt-101 \cite{zhang2020resnest}  & 48.3M  & 10.2  & 82.3  & -        & Focal-Base \cite{yang2021focal}  & 89.8M  & 16.0  & 83.8  & 96.5     \\
ConViT-B \cite{d2021convit}     & 86.5M  & 16.8  & 82.4  & 95.9        & PVTv2-B5 \cite{wang2021pvtv2}    & 82.0M  & 11.8  & 83.8  & 96.6  \\
T2T-ViTt-24 \cite{yuan2021tokens}  & 64.1M  & 15.0  & 82.6  & 95.9           & SE-CoTNetD-152 \cite{li2022contextual} & 55.8M & 17.0 & 84.0 & 97.0 \\
TNT-B  \cite{han2021transformer}       & 65.6M  & 14.1  & 82.9  & 96.3 & LV-ViT-M$^{\dagger}$  \cite{jiang2021all} & 55.8M  & 16.0  & 84.1 & 96.7     \\
DeepViT-L \cite{zhou2021deepvit}    & 58.9M    & 12.5  & 83.1  & -    & VOLO-D2$^{\dagger}$ \cite{yuan2021volo}    & 58.7M  & 14.1  & 85.2  & 97.2  \\
RegionViT-B \cite{chen2021regionvit}  & 72.7M  & 13.0  & 83.2  & 96.1 & VOLO-D3$^{\dagger}$ \cite{yuan2021volo}     & 86.3M  & 20.6  & 85.4  & 97.3  \\
CaiT-S36 \cite{touvron2021going}     & 68.4M    & 13.9  & 83.3  & -   & VOLO-D4$^{\dagger}$ \cite{yuan2021volo}     & 193.0M & 43.8  & 85.7  & 97.4  \\
BoTNet-S1-128 \cite{srinivas2021bottleneck}  & 75.1M & 19.3 & 83.5 & 96.5 & Dual-ViT-L$^{\dagger}$   & 73.0M  & 18.0  & \textbf{85.7}  & \textbf{97.4}  \\
\Xhline{2\arrayrulewidth}
\end{tabular}
\label{table:imagenet}
\vspace{-0.0in}
\end{table*}

\begin{table*}[!tb]
\centering
\caption{Comparison results of our Dual-ViT with other state-of-the-art vision backbones on ImageNet for image recognition at higher resolution (384 $\times$ 384). We split all runs into 3 segments according to the model size (Small, Base, Large). }
\vspace{-0.1in}
\begin{tabular}{c|c|c|c|cc|c|c|c|c|cc}
\Xhline{2\arrayrulewidth}
Network       & Res & Params & GFLOPs & Top-1 & Top-5 & Network      & Res & Params & GFLOPs & Top-1 & Top-5 \\ \hline
\multicolumn{6}{c|} {Small} & \multicolumn{6}{c} {Base} \\ \hline
CvT-13 \cite{wu2021cvt}          & 384 & 20.0M  & 16.3  & 83.0  & 96.4 & CvT-21 \cite{wu2021cvt}             & 384 & 32.0M  & 24.9  & 83.3  & 96.2  \\
T2T-ViT-14 \cite{yuan2021tokens} & 384 & 21.5M  & 17.1  & 83.3  & 96.5  & CrossViT-18 \cite{chen2021crossvit} & 384 & 44.6M  & 32.4   & 83.9  & -     \\
CrossViT-15 \cite{chen2021crossvit} & 384 & 28.5M  & 21.4  & 83.5  & -    & CaiT-S24  \cite{touvron2021going}   & 384 & 46.9M  & 32.2  & 85.1  & 97.4  \\
CaiT-XS24 \cite{touvron2021going}   & 384 & 26.6M  & 19.3  & 84.1  & 96.9 & VOLO-D1 \cite{yuan2021volo}         & 384 & 26.6M  & 22.8  & 85.2  & 97.2 \\
Dual-ViT-S    & 384 & 25.1M  & 18.4  & \textbf{85.2}  & \textbf{97.3}  & Dual-ViT-B  & 384 & 42.6M  & 31.9   & \textbf{86.0}  & \textbf{97.6}  \\ \hline
\multicolumn{12}{c} {Large} \\ \hline
DeiT-B \cite{touvron2021training} & 384 & 86.6M  & 55.4  & 82.9  & 96.2  & CaiT-S36 \cite{touvron2021going} & 384 & 68.2M  & 48.0   & 85.4  & 97.5  \\
CrossViT-18 \cite{chen2021crossvit} & 480 & 44.9M  & 56.6  & 84.1  & -     & LV-ViT-L \cite{jiang2021all} & 448 & 150.5M & 157.2  & 85.9  & -     \\
Swin-B \cite{liu2021swin} & 384 & 88.0M  & 47.1  & 84.5  & 97.0  & VOLO-D2 \cite{yuan2021volo} & 384 & 58.7M  & 46.1   & 86.0  & 97.5  \\
BoTNet-S1-128 \cite{srinivas2021bottleneck} & 384 & 75.1M  & 45.8 & 84.7  & 97.0  & VOLO-D3 \cite{yuan2021volo} & 448 & 86.3M  & 67.9   & 86.3  & 97.7     \\
LV-ViT-M  \cite{jiang2021all} & 384 & 55.8M  & 42.2  & 85.4  & -     & Dual-ViT-L  & 384 & 73.0M  & 60.5   & \textbf{86.5}  & \textbf{97.8}  \\
\Xhline{2\arrayrulewidth}
\end{tabular}
\label{table:imagenetlarge}
\vspace{-0.1in}
\end{table*}

\subsection{Ablation Study of Dual Block}

We firstly study how each particular design in Dual block influences the overall performances of our Dual-ViT backbone for image recognition task. Figure \ref{fig:ablation} shows the detailed architectures of different variants of Dual block and their performances on ImageNet dataset. Here we include three degraded variants of our Dual block by removing (a) the self-attention layer or (b) the feed-forward layer in the semantic pathway, and (c) exchanging the order of self-attention and cross-attention layers in semantic pathway. Note that for fair comparison, we fix the backbone network as Dual-ViT-S, and only the Dual blocks in the first two stages are replaced with each variant.

The first observation when comparing Variant (a) and our complete Dual block (d) is intuitive. The semantic pathway in Variant (a) solely triggers cross-attention learning between input feature and semantic tokens and ignores the inherent dependencies within semantic tokens, thereby performing worse than Dual block (d). In the meanwhile, it is also apparent that when the commonly adopted feed-forward layer is removed in semantic pathway, Variant (b) results in clear performance drops. This confirms that feed-forward layer is a very practical choice to enhance the power of self-attention. One step further is to plug an additional self-attention layer after cross-attention layer in Variant (a). In this way, Variant (c) leads to performance boosts by additionally modeling dependencies among semantic tokens in semantic pathway. However, we notice that such order of self-attention and cross-attention layers in Variant (c) performs worse than Dual block (d) that first employs self-attention among semantic tokens and then conducts cross-attention between input feature and semantic tokens. We speculate that the performance degradations of Variant (c) may be caused by the earlier cross-attention learning that ignores the inherently different peculiarity of input feature and semantic tokens. Instead, the design of earlier self-attention learning in our Dual block (d) enables a specialized encoding process for semantic tokens that modulates them for better cross-attention learning.

\subsection{Performance Comparisons with State-of-the-Art\\ Vision Backbones}

We compare Dual-ViT with various state-of-the-art vision backbones on ImageNet, COCO, and ADE20K datasets for multiple vision tasks (image recognition, object detection, instance segmentation, and semantic segmentation). All runs can be briefly splitted into three segments according to the model size: Small, Base, Large. We also report the model parameters (Params) and GFLOPs to quantify computational cost.

\textbf{Image Recognition.}
Table \ref{table:imagenet} summarizes the quantitative results of our Dual-ViT backbones under different sizes (Dual-ViT-S, Dual-ViT-B, Dual-ViT-L) on ImageNet at typical resolution (254 $\times$ 254). It is also worthy to note that the superior ViTs (e.g., LV-ViT and VOLO) additionally use the advanced strategy of Token Labeling objective via MixToken \cite{jiang2021all}. Similarly, our Dual-ViT$^{\dagger}$ also leverages this strategy to facilitate model learning. In addition, we report the performances of Dual-ViT-S in Small size without Token Labeling objective, enabling fair comparisons against other vision backbones implemented without the advanced strategy.

\begin{table*}[!tb]
\centering
\caption{Comparison results of our Dual-ViT with other state-of-the-art vision backbones on COCO for object detection and instance segmentation downstream tasks. We split all runs into 2 segments according to the model size (Small and Base). The basic detector of RetinaNet is adopted for object detection and we report Average Precision ($AP$) across different IoU thresholds or three different object sizes for evaluation. Mask R-CNN is employed as the basic model for instance segmentation, where both bounding box and mask AP ($AP^b$ and $AP^m$) are reported.}
\begin{tabular}{c|cccccc|cccccc}
\Xhline{2\arrayrulewidth}
\multirow{2}{*}{Backbone} & \multicolumn{6}{c|}{RetinaNet 1x \cite{lin2017focal}}       & \multicolumn{6}{c}{Mask R-CNN 1x \cite{he2017mask}}           \\ \cline{2-13}
   & $AP$ & $AP_{50}$ & $AP_{75}$ & $AP_S$ & $AP_M$ & $AP_L$ & $AP^b$ & $AP^b_{50}$ & $AP^b_{75}$ & $AP^m$  & $AP^m_{50}$ & $AP^m_{75}$ \\ \hline
ResNet50 \cite{he2016deep}                   & 36.3 & 55.3 & 38.6 & 19.3 & 40.0 & 48.8 & 38.0 & 58.6  & 41.4  & 34.4 & 55.1  & 36.7  \\
Swin-T   \cite{liu2021swin}                  & 41.5 & 62.1 & 44.2 & 25.1 & 44.9 & 55.5 & 42.2 & 64.6  & 46.2  & 39.1 & 61.6  & 42.0  \\
Twins-SVT-S \cite{chu2021twins}              & 43.0 & 64.2 & 46.3 & 28.0 & 46.4 & 57.5 & 43.4 & 66.0  & 47.3  & 40.3 & 63.2  & 43.4  \\
RegionViT-S+ \cite{chen2021regionvit}        & 43.9 & -    & -    & -    & -    & -    & 44.2 & -     & -     & 40.8 & -     & -     \\
PVTv2-B2  \cite{wang2021pvtv2}               & 44.6 & 65.6 & 47.6 & 27.4 & 48.8 & 58.6 & 45.3 & 67.1  & 49.6  & 41.2 & 64.2  & 44.4  \\
Dual-ViT-S                                   & \textbf{46.2} & \textbf{67.4} & \textbf{49.9} & \textbf{30.6} & \textbf{49.9} & \textbf{60.9} & \textbf{46.5} & \textbf{68.3}  & \textbf{51.2}  & \textbf{42.2} & \textbf{65.3}  & \textbf{46.1}  \\ \hline \hline
ResNet101 \cite{he2016deep}                  & 38.5 & 57.8 & 41.2 & 21.4 & 42.6 & 51.1 & 40.4 & 61.1  & 44.2  & 40.4 & 61.1  & 44.2  \\
ResNeXt101-64x4d  \cite{xie2017aggregated}   & 41.0 & 60.9 & 44.0 & 23.9 & 45.2 & 54.0 & 42.8 & 63.8  & 47.3  & 38.4 & 60.6  & 41.3 \\
Swin-S \cite{liu2021swin}                    & 44.5 & 65.7 & 47.5 & 27.4 & 48.0 & 59.9 & 44.8 & 66.6  & 48.9  & 40.9 & 63.4  & 44.2  \\
Twins-SVT-B \cite{chu2021twins}              & 45.3 & 66.7 & 48.1 & 28.5 & 48.9 & 60.6 & 45.2 & 67.6  & 49.3  & 41.5 & 64.5  & 44.8  \\
RegionViT-B \cite{chen2021regionvit}         & 44.6 & -    & -    & -    & -    & -    & 45.4 & -     & -     & 41.6 & -     & -     \\
PVTv2-B3  \cite{wang2021pvtv2}               & 45.9 & 66.8 & 49.3 & 28.6 & 49.8 & 61.4 & 47.0 & 68.1  & 51.7  & 42.5 & 65.7  & 45.7  \\
PVTv2-B5  \cite{wang2021pvtv2}               & 46.2 & 67.1 & 49.5 & 28.5 & 50.0 & 62.5 & 47.4 & 68.6  & 51.9  & 42.5 & 65.7  & 46.0  \\
Dual-ViT-B                                   & \textbf{47.4} & \textbf{68.1} & \textbf{51.2} & \textbf{29.6} & \textbf{51.9} & \textbf{63.1} & \textbf{48.4} & \textbf{69.9}  & \textbf{53.3}  & \textbf{43.4} & \textbf{66.7}  & \textbf{46.8}  \\ \Xhline{2\arrayrulewidth}
\end{tabular}
\label{table:od}
\end{table*}

\begin{table*}[!tb]
\centering
\caption{Comparison results of Dual-ViT with other state-of-the-art vision backbones on COCO for object detection downstream task. Here we only include the baselines in small size, and evaluate them under four state-of-the-art object detectors (Cascade Mask R-CNN, ATSS, GFL, and Sparse RCNN). The bounding box AP ($AP^b$) across different IoU thresholds are reported.}
\begin{tabular}{c|ccc|c|ccc}
\Xhline{2\arrayrulewidth}
Backbone           & $AP^b$ & $AP^b_{50}$ & $AP^b_{75}$ & Backbone    & $AP^b$ & $AP^b_{50}$ & $AP^b_{75}$ \\ \hline
\multicolumn{4}{c|}{Cascade Mask R-CNN} & \multicolumn{4}{c}{ATSS}    \\ \hline
ResNet50 \cite{he2016deep} & 46.3 & 64.3  & 50.5  & ResNet50 \cite{he2016deep} & 43.5 & 61.9  & 47.0    \\
Swin-T  \cite{liu2021swin}  & 50.5 & 69.3  & 54.9  & Swin-T  \cite{liu2021swin} & 47.2 & 66.5  & 51.3    \\
PVTv2-B2 \cite{wang2021pvtv2} & 51.1 & 69.8  & 55.3  & PVTv2-B2 \cite{wang2021pvtv2} & 49.9 & 69.1  & 54.1    \\
Dual-ViT-S         & \textbf{52.4} & \textbf{71.0} & \textbf{56.9} & Dual-ViT-S  & \textbf{51.0} & \textbf{69.9} & \textbf{55.9} \\ \hline\hline
\multicolumn{4}{c|}{GFL}  & \multicolumn{4}{c}{Sparse RCNN}                      \\ \hline
ResNet50 \cite{he2016deep} & 44.5 & 63.0  & 48.3  & ResNet50 \cite{he2016deep} & 44.5 & 63.4  & 48.2    \\
Swin-T  \cite{liu2021swin} & 47.6 & 66.8  & 51.7  & Swin-T \cite{liu2021swin} & 47.9 & 67.3  & 52.3    \\
PVTv2-B2 \cite{wang2021pvtv2} & 50.2 & 69.4  & 54.7  & PVTv2-B2 \cite{wang2021pvtv2} & 50.1 & 69.5  & 54.9    \\
Dual-ViT-S         & \textbf{51.3} & \textbf{70.1} & \textbf{55.7} & Dual-ViT-S  & \textbf{51.4} & \textbf{70.9} & \textbf{56.2} \\ \Xhline{2\arrayrulewidth}
\end{tabular}
\label{table:od4}
\end{table*}

\begin{table*}[!tb]
\centering
\caption{Comparison results of our Dual-ViT with other state-of-the-art vision backbones on ADE20K for semantic segmentation downstream task. We split all runs into 2 segments according to the model size (Small and Base). Note that existing works adopt different basic models (e.g., UPerNet, DeeplabV3, and Semantic FPN) for this task. Here we evaluate our Dual-ViT under the most common basic model of UPerNet. Mean IoU (mIoU) averaged over all classes is reported.}
\begin{tabular}{cc|c|cc|c}
\Xhline{2\arrayrulewidth}
\multicolumn{3}{c|} {Small} & \multicolumn{3}{c} {Base} \\ \hline
Method       & Backbone                              & mIoU  & Method       & Backbone    & mIoU \\ \hline
UPerNet \cite{xiao2018unified}     & ResNet-50 \cite{he2016deep}           & 42.8  & UPerNet \cite{xiao2018unified}     & ResNet-101 \cite{he2016deep}             & 44.9 \\
UPerNet \cite{xiao2018unified}     & DeiT-S \cite{touvron2021training}     & 43.8  & UPerNet \cite{xiao2018unified}     & DeiT-B  \cite{touvron2021training}       & 47.2 \\
DeeplabV3 \cite{chen2018encoder}   & ResNeSt-50 \cite{zhang2020resnest}    & 45.1  & DeeplabV3 \cite{chen2018encoder}    & ResNeSt-101 \cite{zhang2020resnest}      & 46.9 \\
Semantic FPN \cite{kirillov2019panoptic} & PVTv2-B2 \cite{wang2021pvtv2}         & 45.2  & Semantic FPN \cite{kirillov2019panoptic} & PVTv2-B3 \cite{wang2021pvtv2}            & 47.3 \\
UPerNet \cite{xiao2018unified}     & RegionViT-S+ \cite{chen2021regionvit}  & 45.3  & UPerNet \cite{xiao2018unified}     & RegionViT-B+ \cite{chen2021regionvit}     & 47.5 \\
UPerNet \cite{xiao2018unified}     & Swin-T \cite{liu2021swin}             & 45.8  & UPerNet \cite{xiao2018unified}     & Twins-SVT-B \cite{chu2021twins}          & 48.9 \\
UPerNet \cite{xiao2018unified}     & Twins-SVT-S \cite{chu2021twins}       & 47.1  & UPerNet \cite{xiao2018unified}     & Swin-S \cite{liu2021swin}                & 49.5 \\
UPerNet \cite{xiao2018unified}     & Dual-ViT-S                            & \textbf{49.2}     & UPerNet \cite{xiao2018unified}     & Dual-ViT-B                               & \textbf{51.7}    \\ \Xhline{2\arrayrulewidth}
\end{tabular}
\label{table:ss}
\vspace{-0.1in}
\end{table*}

In general, it is observed that our Dual-ViT backbones consistently achieve superior performances than state-of-the-art vision backbones with reduced computational cost across most model sizes. Remarkably, under Large size, the top-1 accuracy of Dual-ViT-L$^{\dagger}$ manages to reach 85.7\%, while only requiring 41.1\% GFLOPs and 37.8\% parameters against the superior ViT backbone (VOLO). The results basically show the key merit of decomposing self-attention procedure into global semantics and finer feature extraction procedures in our Dual-ViT, leading to a better trade-off between computational cost and performance. Concretely, Vision Transformer backbones exhibit prominent performances than CNN backbones (ResNet and ResNeXt). Under the same Small size, CrossViT-15 goes beyond DeiT-S (single-branch ViT with fixed patch size) by utilizing two token branches with different patch sizes, leading to performance boosts. We also observe a clear performance gap between CrossViT-15 and RegionViT-S. The reason might be that compared to CrossViT-15 that only fuses the final outputs of two token branches, RegionViT-S triggers more interactions between regional tokens and local tokens throughout the whole architecture via regional-to-local attention. Finally, unlike the local self-attention in RegionViT-S that only mines dependency between single regional token and a subset of local tokens, our Dual-ViT-S holistically exploits the dependency between the global semantics and local features, leading to the best performances.

Next, we further evaluate our Dual-ViT backbones at higher resolution (384 $\times$ 384) under different model sizes (i.e., Dual-ViT-S, Dual-ViT-B, Dual-ViT-L). Table \ref{table:imagenetlarge} details the performances on ImageNet dataset. Note that we follow \cite{yuan2021volo} and fine-tune the pre-trained vision backbones at higher resolution (384 $\times$ 384) with AdamW optimizer \cite{loshchilov2017decoupled} on 8 V100 GPUs. The whole optimization process consists of 35 epochs with cosine decay learning rate scheduler \cite{loshchilov2016sgdr} and 5 epochs of linear warm-up. The learning rate and weight decay are set as $8.0e^{-6}$ and $1.0e^{-8}$. Here we report the input resolution (Res), model parameters (Params) and GFLOPs to quantify computational cost. In general, it is clearly observed that our Dual-ViT backbones consistently achieve superior performances than state-of-the-art vision backbones at higher resolution with reduced computational cost across most model sizes. This again verifies the key advantage of our Dual-ViT that exploits dependency between global semantics and finer pixel level features via an efficient design of two interactive pathways, even under the setup of higher resolution.

\begin{figure*}[!tb]
\centering {\includegraphics[width=0.9\textwidth]{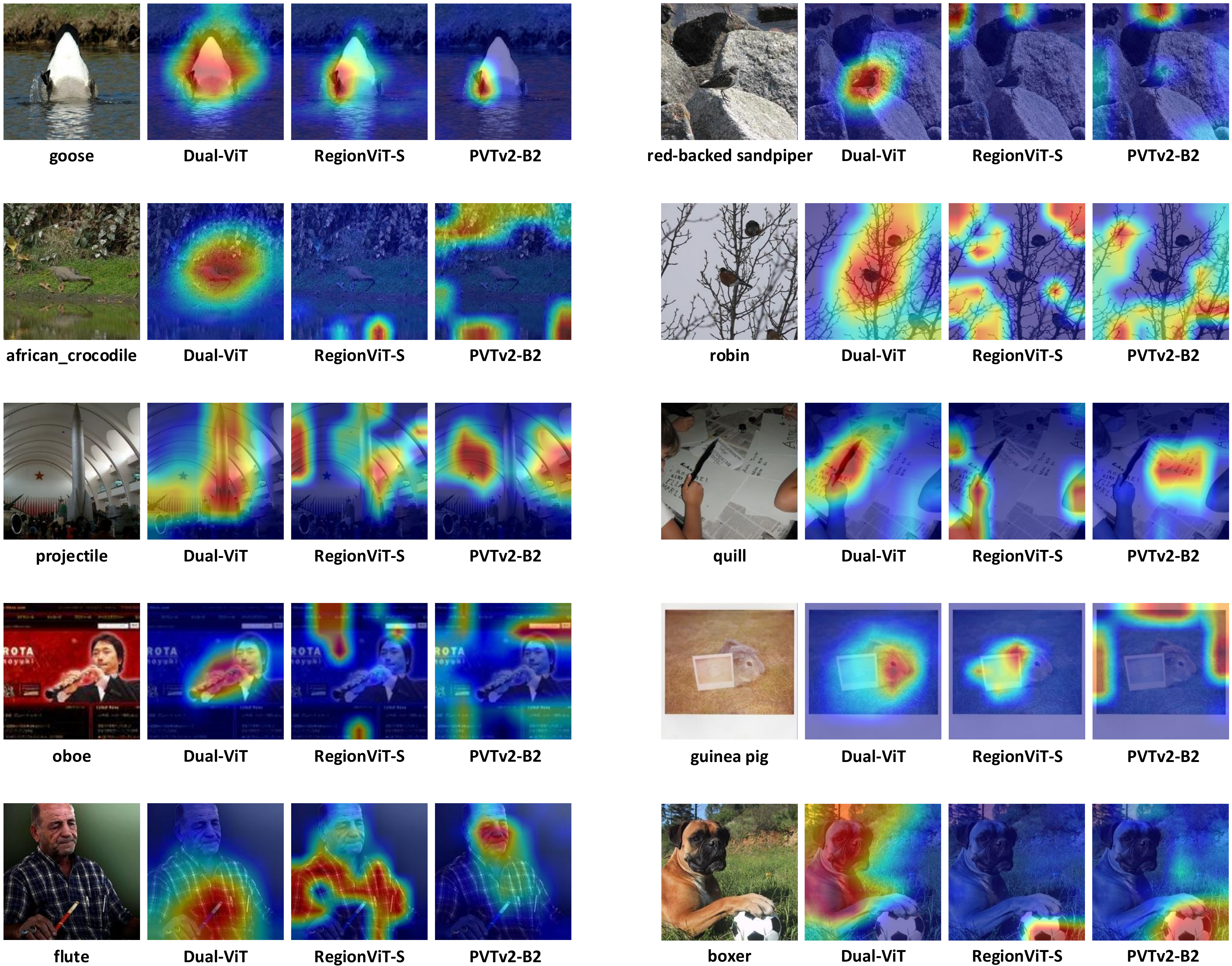}}
\vspace{-0.1in}
\caption{\small Visualization of saliency map produced via Score-CAM \cite{wang2020score} that explains the visual representation learnt by our Dual-ViT, RegionViT-S, and PVTv2-B2 for four images in ImageNet.}
\label{fig:figatt}
\vspace{-0.1in}
\end{figure*}

\textbf{Object Detection and Instance Segmentation.}
To further validate the effectiveness of Dual-ViT as a general vision backbone, we compare with state-of-the-art backbones when transferred to dense prediction downstream tasks of object detection and instance segmentation. Here we first adopt the basic model of RetinaNet and Mask R-CNN for each task, respectively. The performances on COCO are shown in Table \ref{table:od}. Our Dual-ViT-S and Dual-ViT-B apparently have achieved the best performances in terms of all metrics under each model size. Moreover, we follow \cite{wang2021pvtv2} and involve four state-of-the-art object detectors in Table \ref{table:od4} to examine the generalizability of our pre-trained Dual-ViT in object detection downstream task. Generally speaking, we observe similar behaviors of various vision backbones under the four state-of-the-art detectors as that in Table \ref{table:od}. In comparison to PVTv2-B2, Dual-ViT-S still maintains 1.3\%/1.1\%/1.1\%/1.3\% superiority in $AP^b$ under Cascade Mask R-CNN/ATSS/GFL/ Sparse RCNN, proving the unique advantage of enhancing self-attention learning in parallel through coupled pixel and semantic pathways.

\textbf{Semantic Segmentation.}
Table \ref{table:ss} lists the performance comparisons on ADE20K for semantic segmentation downstream task. Similarly, under the same basic architecture of UPerNet, Dual-ViT-S/Dual-ViT-B performs more than 2.1\%/2.2\% better than the best competitor Twins-SVT-S/Swin-S, respectively. Such phenomenon verifies that it is always beneficial to incorporate a critical semantic pathway in Multi-scale ViT architecture, that efficiently compresses token vectors into global semantics and meanwhile enhances self-attention learning with the interactions between two pathways.

\subsection{Visualization Analysis}
Finally, to better qualitatively examine the quality of learnt visual representation via our Dual-ViT, we visualize the saliency map produced by Score-CAM \cite{wang2020score} for visual representations learnt by Dual-ViT-S, RegionViT-S, and PVTv2-B2 in Figure \ref{fig:figatt}. For each input image, the saliency map of learnt representation indicates the importance of each pixel in reflecting the class discrimination. As shown in this figure, it is intriguing to see that compared to RegionViT-S and PVTv2-B2, Dual-ViT-S captures more meaningful pixels in saliency map that benefit the recognition of target object. This again confirms that the capacity of learnt visual representation by our Dual-ViT are stronger.

\section{Conclusions}
In this work, we present Dual Vision Transformer (Dual-ViT), a new multi-scale ViT backbone which novelly models self-attention learning in two interactive pathways: the pixel pathway for learning finer pixel level details and the semantic pathway that extracts holistic global semantic information from inputs. The learnt semantic tokens from semantic pathway further serve as high-level semantic prior to facilitate the finer local feature extraction in pixel pathway. In this way, the enhanced self-attention information is spread in parallel along the two pathways, pursuing a better accuracy-latency trade-off. Extensive empirical results on various vision tasks demonstrate the superiority of Dual-ViT against state-of-the-art ViTs.


%

%
%
%
%
%
%
%
%
%
%
%

\bibliographystyle{IEEEtran}
\bibliography{egbib}

\end{document}